\documentclass[10pt,twocolumn,letterpaper]{article}

\usepackage{cvpr}
\usepackage{times}
\usepackage{epsfig}
\usepackage{graphicx}
\usepackage{amsmath}
\usepackage{amssymb}
\usepackage{xcolor}
\usepackage{pdfpages}
\usepackage{tikz}
\usepackage{pgfplots}
\usepackage{amsmath}
\pgfplotsset{compat=1.9}
\usepgfplotslibrary{colorbrewer}
\pgfplotsset{cycle list/Dark2}

\usepackage[breaklinks=true,bookmarks=false]{hyperref}

\cvprfinalcopy 

\def\cvprPaperID{****} 

\begin{document}

\title{Semi-Supervised Exploration in Image Retrieval}

\author{
Cheng Chang~\thanks{Authors contributed equally to this work.}\\
Layer6 AI\\
{\tt\small jason@layer6.ai}
\and
Himanshu Rai~\footnotemark[1]\\
Layer6 AI\\
{\tt\small himanshu@layer6.ai}
\and
Satya Krishna Gorti~\footnotemark[1]\\
Layer6 AI\\
{\tt\small satya@layer6.ai}
\and
Junwei Ma~\footnotemark[1]\\
Layer6 AI\\
{\tt\small jeremy@layer6.ai}
\and
Chundi Liu~\footnotemark[1]\\
Layer6 AI\\
{\tt\small chundi@layer6.ai}
\and
Guangwei Yu\\
Layer6 AI\\
{\tt\small guang@layer6.ai}
\and
Maksims Volkovs\\
Layer6 AI\\
{\tt\small maks@layer6.ai}
}

\maketitle

\begin{abstract}
   We present our solution to Landmark Image Retrieval Challenge 2019. This challenge was based on the large Google Landmarks Dataset V2\cite{DBLP:journals/corr/abs-1812-01584}. The goal was to retrieve all database images containing the same landmark for every provided query image. Our solution is a combination of global and local models to form an initial KNN graph. We then use a novel extension of the recently proposed graph traversal method EGT \cite{egt} referred to as semi-supervised EGT to refine the graph and retrieve better candidates.
\end{abstract}

\section{Approach}
\paragraph{Introduction}
Image retrieval is a fundamental problem in computer
vision, where the goal is to rank relevant images from an
index set given a query image. Landmark retrieval in
particular is an important task as people often take
photographs that contain landmarks. The Google Landmark
Retrieval 2019 challenge aims to advance research on this
task, introducing the largest worldwide dataset to date and
providing a standardized framework for benchmark.
The challenge involves retrieving the top 100
candidates from an index set of 700K images for each of the
100K query in the test set. A training set
of 4.1 million images with over 200K unique landmarks,
which we call Train-V2. Additionally, we use the Google
Landmarks V1\cite{Noh2017LargeScaleIR} train set with
1 million images and 30K unique landmarks, which we call
Train-V1.

\begin{figure}
    \includegraphics[width=0.50\textwidth]{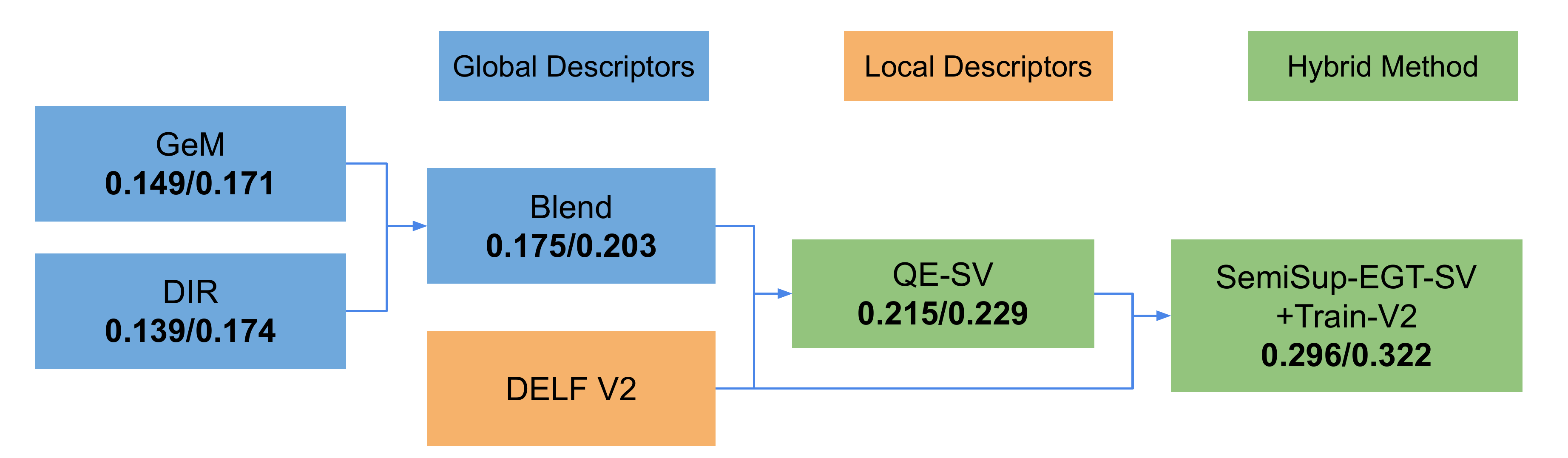}
  \caption{Overview of our model pipeline: two CNN modes (GEM and DIR) are used to extract global descriptors. We concatenate them to get the blended global descriptors. Local descriptors were extracted from DELF-V2 model. Then a query expansion method with spatial verification (denoted as QE-SV) is applied to obtain a better KNN graph. To leverage label information from Train-V2, we propose semi-supervised EGT (referred as SemiSup-EGT). }\label{fig:pipe}
\end{figure}

Figure~\ref{fig:pipe} outlines our pipeline. Global CNN
descriptors generate inner product distance used to build a
$k$-nearest neighbor (KNN) graph $G_k$. We then apply the 
recently proposed EGT algorithm to further improve retrieval.
EGT builds \emph{trusted} paths on $G_k$, alternating between
exploring neighbors and exploiting most confident edges.
Starting with query as the only  trusted vertex, the explore step
adds neighbors of trusted vertices to a priority queue ordered
by edge weights. The exploit step then retrieves all vertices
that have traversed edge weights larger than a threshold $t$,
called trusted vertices. The path formed is referred to as
the trusted path, and the explore/exploit steps are repeated.
The motivation is that relevant images may be visually dissimilar,
but share a similar image that can ``bridge'' the gap.
However this approach can fail when no such image exists in the
index, limiting exploration. In order to overcome this, we
propose the semi-supervised EGT to expand exploration.

\begin{figure}
    \includegraphics[width=0.50\textwidth]{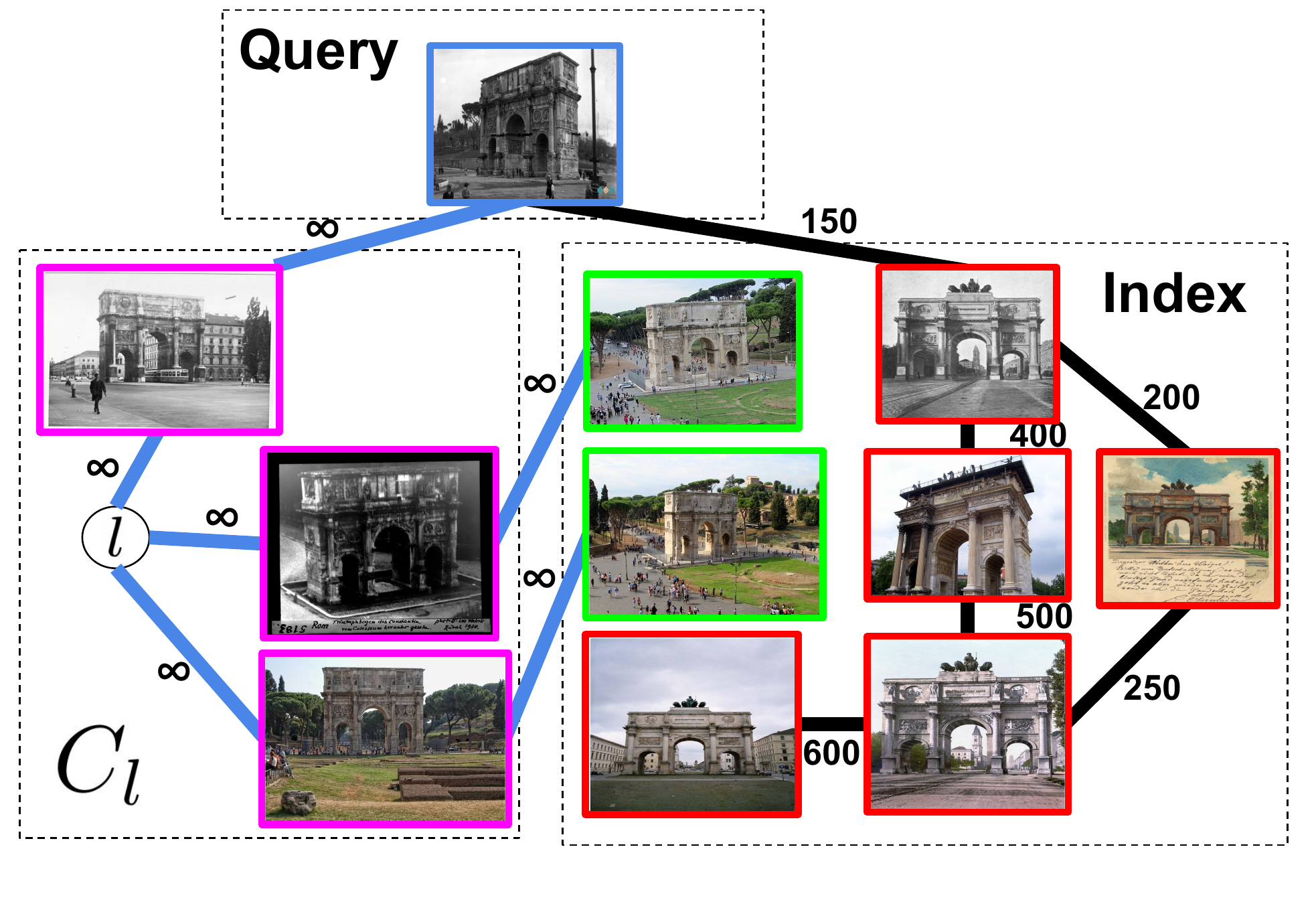}
  \caption{Example of semi-supervised EGT on the Google Landmark Challenge 2019 datasets. Query and Train-V2 are highlighted in blue and purple respectively. For index image set, relevant images are highlighted in green while irrelevant ones are in red. The trusted paths introduced by semi-supervised EGT are highlighted in blue. Best viewed in color.}\label{fig:semiEGT}
\end{figure}

\paragraph{Semi-Supervised EGT}
The main idea of our approach is to leverage label information
from a Train-V2. For each label $l\in\{1,2,...,L\}$, corresponding
images in Train-V2 are connected to it, resulting in the set of sub-graphs
$\{C_l\}$. The edges are set to maximum weights to indicate highest level of
similarities. An example for a particular $C_l$ is shown in the
left part of Figure~\ref{fig:semiEGT}. During retrieval, one of the
sub-graph $C_l$ is then added to $G_k$, potentially introducing
new paths between query and index images. An advantage of this
approach is that paths between queries and indices can be formed
using the the corresponding $C_l$ as bridges, regardless of their
visual similarities.

In order to connect query to $C_l$, we calculate the pair-wise
similarity scores between the query and each images in Train-V2. Then,
a majority vote on the top-3 candidates' labels select a particular label
$l$ for the query. In the event of a tie, no label is selected and we proceed without using Train-V2, otherwise the most similar image in Train-V2 with
label $l$ is connected to the query with maximum edge weight.
This process is repeated for each of the index images.
Figure~\ref{fig:semiEGT} shows an example
of this algorithm. The black path on the right depict the original EGT graph,
and the blue path on the left depict the new paths introduced through $C_l$.
Two relevant images in the index set, shown in green, shares the same $C_l$ as
query, and are thus retrieved.

Constructing the graph this way ensures that the connected sub-graph $C_l$ is given precedence when building the trusted paths by EGT, consistent with the motivation of EGT, where more trustworthy neighbors are explored with higher priority. Once the graph is built, we apply EGT, skipping retrieval for
the Train-V2 images as they are not part of the index set to be retrieved.

\section{Experiments}

Figure~\ref{fig:pipe} depicts our pipeline. We use two 
global descriptor models GeM~\cite{Radenovic2018} and
DIR~\cite{gordo2016deep}. The GeM model is trained
with a ResNet-101 backbone\cite{He2016DeepRL} pre-trained on ImageNet \cite{imagenet_cvpr09} with GeM pooling and a fully connected whitening layer
as described in \cite{Radenovic2018}. All the trainable parameters are fine-tuned on Train-V1 dataset.
We concatenated the GeM and DIR vectors to obtain our global descriptors referred to as the blend model. The top k (k=100) candidates ranked by their inner product with query based on the blend model form our initial KNN graph.

We use DELF-V2~\cite{DBLP:journals/corr/abs-1812-01584} as local descriptors, applying
RANSAC-based spatial verification~\cite{Fischler:1981:RSC:358669.358692} (SV) to re-rank top 10 index image candidates for each query. Among the re-ranked
candidates, we use two most reliable index images to perform
query expansion (QE)~\cite{qe}. This process is extended to database
side by issuing every index image as query. This is followed by our 
semi-supervised EGT approach that further refines the graph and improves
the retrieval mAP.
\paragraph{Results}
\begin{table}[h!]
\centering
\begin{tabular}{ |p{4.9cm}||p{1.1cm}|p{1.1cm}|  }
 \hline
 \multicolumn{3}{|c|}{Result (mAP)} \\
 \hline
 Method & Public & Private\\
 \hline
 Blend     & 0.1753 &   0.2030\\
 Blend+QE-SV & 0.2150   & 0.2290\\
 Blend+QE-SV+EGT & 0.2545   & 0.2672\\
 Blend+QE-SV+SemiSup-EGT  & 0.2964 &  0.3218\\
 \hline
\end{tabular}
\caption{Ablation of the proposed pipeline on public and private leaderboard. }
\label{table:1}
\end{table}

We present our pipeline's results on public and private set as shown in Table \ref{table:1}.  Our baseline is concatenation of GeM and DIR embedding (blend) followed by inner product retrieval. Applying QE on spatially verified retrieved images (QE-SV) improves performance, and further applying EGT to the pipeline gives another significant improvement. EGT bridges query and index images that share visual similarity with other images but are otherwise dissimilar based on global or local descriptors.

With the proposed semi-supervised EGT, we achieve a further 3.5 point improvement on the public leaderboard and over 5 point improvement on the private leaderboard.
The proposed semi-supervised EGT extends EGT to leverage additional labeled data outside the index set by expanding on the idea of traversing trusted edges in EGT. Shared label information between the Train-V2 and index set images
form additional bridges between visually dissimilar images which is difficult in the original unsupervised version of EGT that rely on global and local descriptors.

\paragraph{Conclusion}
In this paper, we describe our approach for the 2019 Landmark Retrieval Challenge. We present a novel semi-supervised approach that extends EGT when an additional labeled dataset is available. The model achieves very competitive score on the challenge without relying on data cleaning methods.

{\small
\bibliographystyle{ieee}
\bibliography{challenge}
}

\end{document}